\documentclass[runningheads]{llncs}
\usepackage{graphicx}
\usepackage{amsmath,amssymb} 
\usepackage{color}
\usepackage{epsfig}
\usepackage{booktabs}
\usepackage{soul} 

\usepackage{subfig} 

\usepackage[width=122mm,left=12mm,paperwidth=146mm,height=193mm,top=12mm,paperheight=217mm]{geometry}

\begin{document}
\pagestyle{headings}
\mainmatter
\title{Temporal Relational Reasoning in Videos} 

\titlerunning{Temporal Relational Reasoning in Videos}

\authorrunning{B. Zhou, A. Andonian, A. Oliva, and A. Torralba}

\author{Bolei Zhou, Alex Andonian, Aude Oliva, Antonio Torralba}


\institute{MIT CSAIL\\
	\email{ \{bzhou,aandonia,oliva,torralba\}@csail.mit.edu}
}

\maketitle

\begin{abstract}
Temporal relational reasoning, the ability to link meaningful transformations of objects or entities over time, is a fundamental property of intelligent species. In this paper, we introduce an effective and interpretable network module, the Temporal Relation Network (TRN), designed to learn and reason about temporal dependencies between video frames at multiple time scales. We evaluate TRN-equipped networks on activity recognition tasks using three recent video datasets - Something-Something, Jester, and Charades - which fundamentally depend on temporal relational reasoning. Our results demonstrate that the proposed TRN gives convolutional neural networks a remarkable capacity to discover temporal relations in videos. Through only sparsely sampled video frames, TRN-equipped networks can accurately predict human-object interactions in the Something-Something dataset and identify various human gestures on the Jester dataset with very competitive performance. TRN-equipped networks also outperform two-stream networks and 3D convolution networks in recognizing daily activities in the Charades dataset. Further analyses show that the models learn intuitive and interpretable visual common sense knowledge in videos\footnote{Code and models are available at \url{http://relation.csail.mit.edu/}.}.


\end{abstract}

\section{Introduction}

\begin{figure}
\includegraphics[width=1\linewidth]{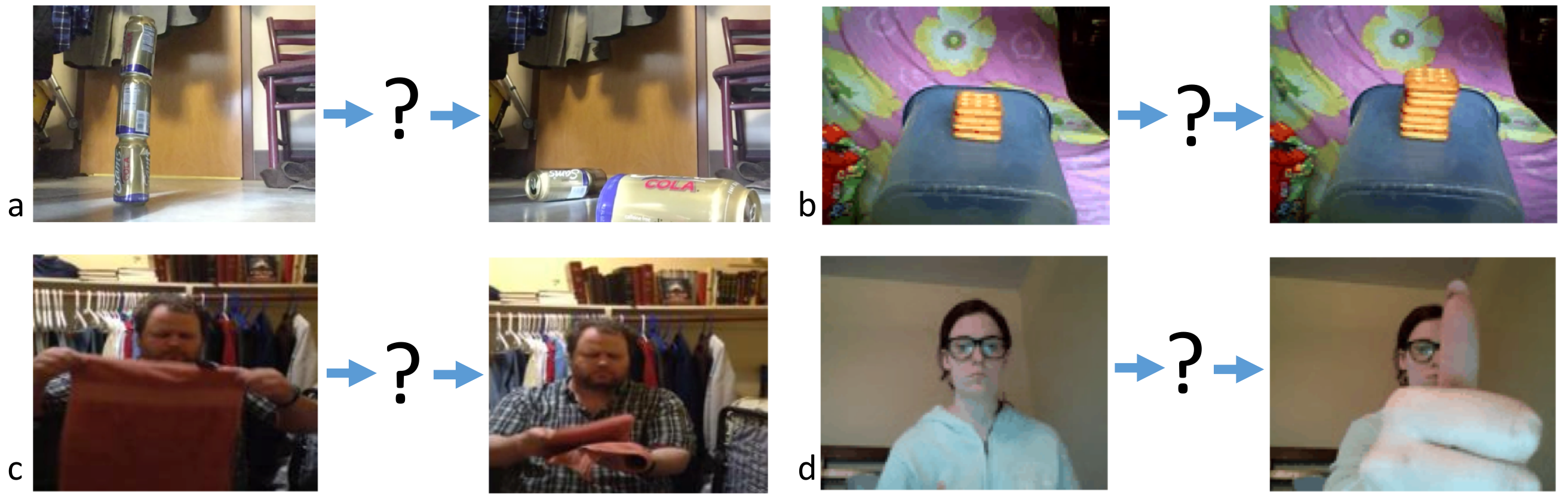}
\caption{What takes place between two observations? (see answer below the first page). Humans can easily infer the temporal relations and transformations between these observations, but this task remains difficult for neural networks.}
\label{cover}
\vspace{-5mm}
\end{figure}

The ability to reason about the relations between entities over time is crucial for intelligent decision-making. Temporal relational reasoning allows intelligent species to analyze the current situation relative to the past and formulate hypotheses on what may happen next. For example (Fig.\ref{cover}), given two observations of an event, people can easily recognize the temporal relation between two states of the visual world and deduce what has happened between the two frames of a video\footnote{Answer: a) Poking a stack of cans so it collapses; b) Stack something; c) Tidying up a closet; d) Thumb up.}.

Temporal relational reasoning is critical for activity recognition, forming the building blocks for describing the steps of an event. A single activity can consist of several temporal relations at both short-term and long-term timescales. For example, the activity of \textit{sprinting} contains the long-term temporal relations of crouching at the starting blocks, running on track, and finishing at the end line, while it also includes the short-term temporal relations of periodic hands and feet movement.

Activity recognition in videos has been one of the core topics in computer vision. However, it remains difficult due to the ambiguity of describing activities at appropriate timescales \cite{sigurdsson2017actions}. Many video datasets, such as UCF101 \cite{soomro2012ucf101}, Sport1M \cite{karpathy2014large}, and THUMOS \cite{gorban2015thumos}, include many activities that can be recognized without reasoning about the long-term temporal relations: still frames and optical flow are sufficient to identify many of the labeled activities. Indeed, the classical two-stream Convolutional Neural Network \cite{simonyan2014two} and the recent I3D Network \cite{carreira2017quo}, both based on frames and optical flow, perform activity recognition very well on these datasets. 

However, convolutional neural networks still struggle in situations where data and observations are limited, or where the underlying structure is characterized by transformations and temporal relations, rather than the appearance of certain entities \cite{santoro2017simple,lake2016building}. It remains remarkably challenging for convolutional neural networks to reason about temporal relations and to anticipate what transformations are happening to the observations. Fig.\ref{cover} shows such examples. The networks are required to discover visual common sense knowledge over time beyond the appearance of objects in the frames and the optical flow.

In this work, we propose a simple and interpretable network module called Temporal Relation Network (TRN) that enables temporal relational reasoning in neural networks. This module is inspired by the relational network proposed in \cite{santoro2017simple}, but instead of modeling the spatial relations, TRN aims to describe the temporal relations between observations in videos. Thus, TRN can learn and discover possible temporal relations at multiple time scales. TRN is a general and extensible module that can be used in a plug-and-play fashion with any existing CNN architecture. We apply TRN-equipped networks on three recent video datasets (Something-Something \cite{goyal2017something}, Jester \cite{jester}, and Charades \cite{sigurdsson2016hollywood}), which are constructed for recognizing different types of activities such as human-object interactions and hand gestures, but all depend on temporal relational reasoning. The TRN-equipped networks achieve very competitive results even given only discrete RGB frames, bringing significant improvements over baselines. Thus TRN provides a practical solution for standard neural networks to solve activity recognition tasks using temporal relational reasoning.

\subsection{Related Work}

\textbf{Convolutional Neural Networks for Activity Recognition}. Activity recognition in videos is a core problem in computer vision. With the rise of deep convolutional neural networks (CNNs) which achieve state-of-the-art performance on image recognition tasks \cite{krizhevsky2012imagenet,zhou2014learning}, many works have looked into designing effective deep convolutional neural networks for activity recognition \cite{karpathy2014large,simonyan2014two,donahue2015long,tran2015learning,wang2016temporal,carreira2017quo}. For instance, various approaches of fusing RGB frames over the temporal dimension are explored on the Sport1M dataset \cite{karpathy2014large}. Two stream CNNs with one stream of static images and the other stream of optical flows are proposed to fuse the information of object appearance and short-term motions \cite{simonyan2014two}. 3D convolutional networks \cite{tran2015learning} use 3D convolution kernels to extract features from a sequence of dense RGB frames. Temporal Segment Networks sample frames and optical flow on different time segments to extract information for activity recognition \cite{wang2016temporal}. A CNN+LSTM model, which uses a CNN to extract frame features and an LSTM to integrate features over time, is also used to recognize activities in videos \cite{donahue2015long}. Recently, I3D networks \cite{carreira2017quo} use two stream CNNs with inflated 3D convolutions on both dense RGB and optical flow sequences to achieve state of the art performance on the Kinetics dataset \cite{kay2017kinetics}. There are several important issues with existing CNNs for action recognition: 1) The dependency on beforehand extraction of optical flow lowers the efficiency of the recognition system; 2) The 3D convolutions on sequences of dense frames are computationally expensive, given the redundancy in consecutive frames; 3) Since sequences of frames fed into the network are usually limited to 20 to 30 frames, it is difficult for the networks to learn long-term temporal relations among frames. To address these issues, the proposed Temporal Relation Network sparsely samples individual frames and then learns their causal relations, which is much more efficient than sampling dense frames and convolving them. We show that TRN-equipped networks can efficiently capture temporal relations at multiple time scales and outperform dense frame-based networks using only sparsely sampled video frames.

\textbf{Temporal Information in Activity Recognition}. For activity recognition on many existing video datasets such as UCF101 \cite{soomro2012ucf101}, Sport1M \cite{karpathy2014large}, THUMOS \cite{gorban2015thumos}, and Kinetics \cite{kay2017kinetics}, the appearance of still frames and short-term motion such as optical flow are the most important information to identify the activities. Thus, activity recognition networks such as Two Stream network \cite{simonyan2014two} and the I3D network \cite{carreira2017quo} are tailored to capture these short-term dynamics of dense frames. Therefore, existing networks don't need to build temporal relational reasoning abilities. On the other hand, recently there have been various video datasets collected via crowd-sourcing, which focus on sequential activity recognition: Something-Something dataset \cite{goyal2017something} is collected for generic human-object interaction. It has video classes such as `Dropping something into something', `Pushing something with something', and even `Pretending to open something without actually opening it'. Jester dataset \cite{jester} is another recent video dataset for gesture recognition. Videos are recorded by crowd-source workers performing 27 kinds of gestures such as `Thumbing up', `Swiping Left', and `Turning hand counterclockwise'. Charades dataset is also a high-level human activity dataset that collects videos by asking crowd workers to perform a series of home activities and then record themselves \cite{sigurdsson2016hollywood}. For recognizing the complex activities in these three datasets, it is crucial to integrate temporal relational reasoning into the networks. Besides, many previous works model the temporal structures of videos for action recognition and detection using bag of words, motion atoms, or action grammar \cite{gaidon2013temporal,pirsiavash2014parsing,wang2013action,gaidon2014activity,wang2016mofap}. Instead of designing temporal structures manually, we use a more generic structure to learn the temporal relations in end-to-end training. One relevant work on modeling the cause-effect in videos is \cite{wang2016actions}. \cite{wang2016actions} uses a two-stream siamese network to learn the transformation matrix between two frames, then uses brute force search to infer the action category. Thus the computation cost is high. Our TRN much more efficiently integrates the multiple frames’ information, both in training and testing.

\textbf{Relational Reasoning and Intuitive Physics}. Recently, relational reasoning module has been proposed for visual question answering with super-human performance \cite{santoro2017simple}. Our work is inspired by that work, but we focus on modeling the multi-scale temporal relations in videos. In the domain of robot self-supervised learning, many models have been proposed to learn the intuitive physics among frames. Given an initial state and a goal state, the inverse dynamics model with reinforcement learning is used to infer the transformation between the object states \cite{agrawal2016learning}. Physical interaction and observations are also used to train deep neural networks \cite{pinto2016curious}. Time contrast networks are used for self-supervised imitation learning of object manipulation from third-person video observation \cite{sermanet2017time}. Our work aims to learn various temporal relations in videos in a supervised learning setting. The proposed TRN can be extended to self-supervised learning for robot object manipulation. 

\begin{figure*}
\vspace{-7mm}
\includegraphics[width=1\linewidth]{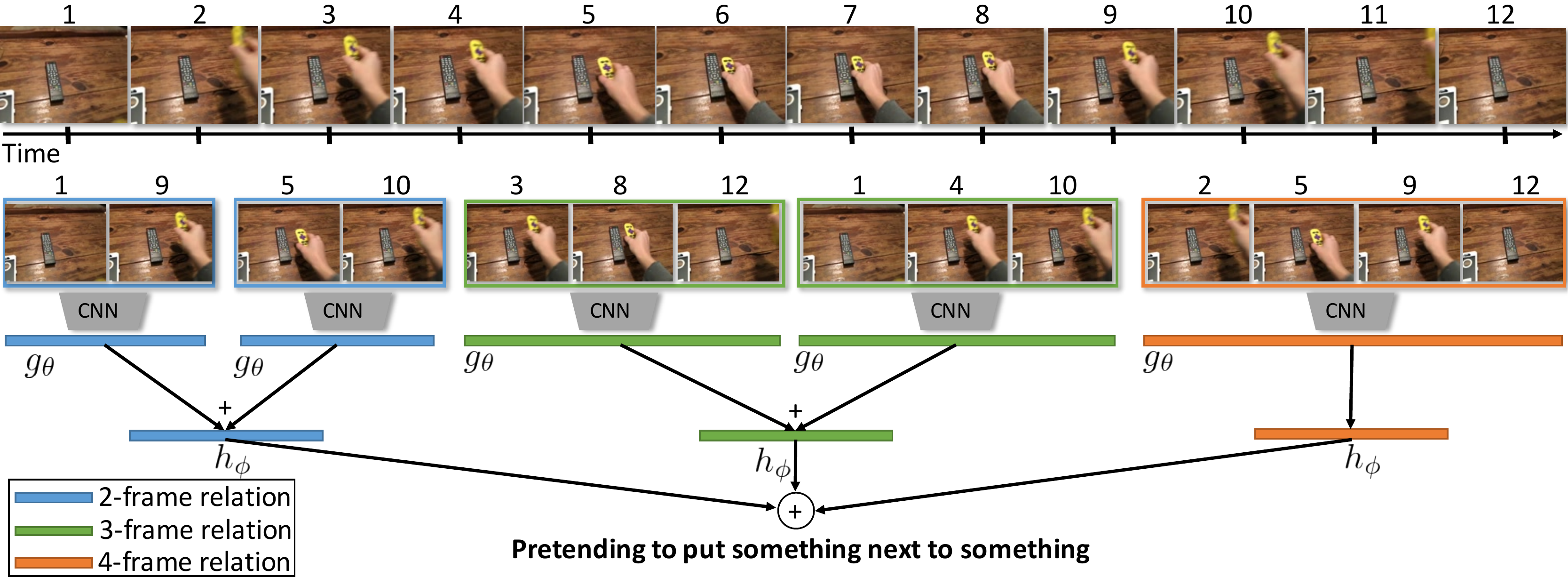}
\vspace{-5mm}
\caption{The illustration of Temporal Relation Networks. Representative frames of a video (shown above) are sampled and fed into different frame relation modules. Only a subset of the 2-frame, 3-frame, and 4-frame relations are shown, as there are higher frame relations included.}
\label{illustration}
\vspace{-8mm}
\end{figure*}

\section{Temporal Relation Networks}

In this section, we introduce the framework of Temporal Relation Networks. It is simple and can be easily plugged into any existing convolutional neural network architecture to enable temporal relational reasoning. In later experiments, we show that TRN-equipped networks discover interpretable visual common sense knowledge to recognize activities in videos.

\subsection{Defining Temporal Relations}

Inspired by the relational reasoning module for visual question answering \cite{santoro2017simple}, we define the pairwise temporal relation as a composite function below: 
\begin{align}
T_2(V) = h_{\phi}\Big( \sum_{i < j}g_\theta(f_{i}, f_{j})\Big)
\end{align}
where the input is the video $V$ with $n$ selected ordered frames as $V=\{f_1,f_2,...,f_n\}$, where $f_i$ is a representation of the $i^{th}$ frame of the video, e.g., the output activation from some standard CNN. The functions $h_{\phi}$ and $g_\theta$ fuse features of different ordered frames. Here we simply use multilayer perceptrons (MLP) with parameters $\phi$ and $\theta$ respectively.  For efficient computation, rather than adding all the combination pairs, we uniformly sample frames $i$ and $j$ and sort each pair.

We further extend the composite function of the 2-frame temporal relations to higher frame relations such as the 3-frame relation function below: 
\begin{align}
T_{3}(V) = h_{\phi}^{'}\Big( \sum_{i < j < k}g_\theta^{'}(f_{i}, f_{j}, f_{k})\Big)
\end{align}
where the sum is again over sets of frames $i, j, k$  that have been uniformly sampled and sorted.

\subsection{Multi-Scale Temporal Relations}
To capture temporal relations at multiple time scales, we use the following composite function to accumulate frame relations at different scales:
\begin{align}
MT_{N}(V) = T_{2}(V) + T_{3}(V) ... + T_{N}(V)
\end{align}
Each relation term $T_d$ captures temporal relationships between $d$ ordered frames.  Each $T_d$ has its own separate $h_{\phi}^{(d)}$ and $g_\theta^{(d)}$. Notice that for any given sample of $d$ frames for each $T_d$, all the temporal relation functions are end-to-end differentiable, so they can all be trained together with the base CNN used to extract features for each video frame. The overall network framework is illustrated in Fig.\ref{illustration}.

\subsection{Efficient Training and Testing}

When training a multi-scale temporal network, we could sample the sums by selecting different sets of $d$ frames for each $T_d$ term for a video. However, we use a sampling scheme that reduces computation significantly. First, we uniformly sample a set of $N$ frames from the N segments of the video, $V^*_N \subset V$, and we use $V^*_N$ to calculate $T_N(V)$.  Then, for each $d<N$, we choose $k$ random subsamples of $d$ frames $V^*_{kd} \subset V^*_N$.  These are used to compute the $d$-frame relations for each $T_{d}(V)$.  This allows $kN$ temporal relations to be sampled while run the base CNN on only $N$ frames, while all the parts are end-to-end trained together.  

At testing time, we can combine the TRN-equipped network with a queue to process streaming video very efficiently. A queue is used to cache the extracted CNN features of the equidistant frames sampled from the video, then those features are further combined into different relation tuples which are further summed up to predict the activity. The CNN feature is extracted from incoming key frame only once then enqueued, thus TRN-equipped networks is able to run in real-time on a desktop to processing streaming video from a webcam.

\section{Experiments}

We evaluate the TRN-equipped networks on a variety of activity recognition tasks. For recognizing activities that depend on temporal relational reasoning, TRN-equipped networks outperform a baseline network without a TRN by a large margin. We achieve highly competitive results on the Something-Something dataset for human-interaction recognition \cite{goyal2017something} and on the Jester dataset for hand gesture recognition \cite{jester}. The TRN-equipped networks also obtain competitive results on activity classification in the Charades dataset \cite{sigurdsson2016hollywood}, outperforming the Flow+RGB ensemble models \cite{sigurdsson2016asynchronous,sigurdsson2016hollywood} using only sparsely sampled RGB frames. 

The statistics of the three datasets Something-Something dataset (Something-V1 \cite{goyal2017something} and Something-V2 \cite{mahdisoltani2018fine} where the Something-V2 is the 2nd release of the dataset in early July 2018) \cite{goyal2017something,mahdisoltani2018fine}, Jester dataset \cite{jester}, and Charades dataset \cite{sigurdsson2016hollywood} are listed in Table \ref{datasets}. All three datasets are crowd-sourced, in which the videos are collected by asking the crowd-source workers to record themselves performing instructed activities. Unlike the Youtube-type videos in UCF101 and Kinetics, there is usually a clear start and end of each activity in the crowd-sourced video, emphasizing the importance of temporal relational reasoning.

\vspace{-8mm}
\begin{table}\caption{Statistics of the datasets used in evaluating the TRNs.}
\label{datasets}
\small
\centering
\begin{tabular}{p{2.4cm} p{1.5cm}  p{1.5cm}  p{4cm} }
\toprule      
Dataset & Classes & Videos & Type \\ 
\hline  
Something-V1 & 174 & 108,499 & human-object interaction \\ 
Something-V2 & 174 & 220,847 & human-object interaction  \\ 
Jester & 27 & 148,092 & human hand gesture \\
Charades & 157 & 9,848 & daily indoor activity \\
\bottomrule
\end{tabular}
\vspace{-10mm}
\end{table}


\subsection{Network Architectures and Training}

The networks used for extracting image features play an important factor in visual recognition tasks \cite{sharif2014cnn}. Features from deeper networks such as ResNet \cite{he2016deep} usually perform better. Our goal here is to evaluate the effectiveness of the TRN module for temporal relational reasoning in videos. Thus, we fix the base network architecture to be the same throughout all the experiments and compare the performance of the CNN model with and without the proposed TRN modules. 

We adopt Inception with Batch Normalization (BN-Inception) pretrained on ImageNet used in \cite{ioffe2015batch} because of its balance between accuracy and efficiency. We follow the training strategies of partial BN (freezing all the batch normalization layers except the first one) and dropout after global pooling as used in \cite{wang2016temporal}. We keep the network architecture of the MultiScale TRN module and the training hyper-parameters the same for training models on all the three datasets. We set $k=3$ in the experiments as the number of accumulated relation triples in each relation module. $g_{\phi}$ is simply a two-layer MLP with 256 units per layer, while $h_{\phi}$ is a one-layer MLP with the unit number matching the class number. The CNN features for a given frame is the activation from the BN-Inception's global average pooling layer (before the final classification layer). Given the BN-Inception as the base CNN, the training can be finished in less than 24 hours for 100 training epochs on a single Nvidia Titan Xp GPU. In the Multi-Scale TRN, we include all the TRN modules from 2-frame TRN up to 8-frame TRN (thus $N=8$ in Eq.3), as including higher frame TRNs brings marginal improvement and lowers the efficiency.

\subsection{Results on Something-Something Dataset}

Something-Something is a recent video dataset for human-object interaction recognition. There are 174 classes, some of the ambiguous activity categories are challenging, such as `Tearing Something into two pieces' versus `Tearing Something just a little bit', `Turn something upside down' versus `Pretending to turn something upside down'. We can see that the temporal relations and transformations of objects rather than the appearance of the objects characterize the activities in the dataset.

The results on the validation set and test set of Something-V1 and Something-V2 datasets are listed in Table \ref{fig:something_result}a. The baseline is the base network trained on single frames randomly selected from each video. Networks with TRNs outperform the single frame baseline by a large margin. We construct the 2-stream TRN by simply averaging the predicted probabilities from the the two streams for any given video). The 2-stream TRN further improves the accuracy on the validation set of Something-v1 and Something-v2 to \textbf{42.01\%} and \textbf{55.52\%} respectively. Note that we found that the optical stream with average pooling of frames used in TSN \cite{wang2016temporal} achieves better score than the one with the proposed temporal relational pooling so we use 8-frame TSN on optical flow stream, which gets 31.63\% and 46.41\% on the validation set of Something-V1 and Something-V2 respectively. We further submit MultiScale TRN and 2-stream TRN predictions on the test set, the results are shown in Table \ref{fig:something_result}.a 

We compare the TRN with TSN \cite{wang2016temporal}, to verify the importance of temporal orders. Instead of concatenating the features of temporal frames, TSN simply averages the deep features so that the model only captures the co-occurrence rather than the temporal ordering of patterns in the features. We keep all the training conditions the same, and vary the number of frames used by two models. As shown in Table \ref{fig:something_result}b, our models outperform TSNs by a large margin. This result shows the importance of frame order for temporal relation reasoning. We also see that additional frames included in the relation bring further significant improvements to TRN.



\begin{table}%
\vspace{-5mm}
    \centering
    \subfloat[]{
      
       \begin{tabular}{ p{2.5cm}  c c | c c }
      \toprule       
      & \multicolumn{2}{c}{Something-V1} & \multicolumn{2}{c}{Something-V2} \\
       & Val & Test & Val & Test \\
       \hline 
      Baseline & 11.41 & - & - & - \\
      MultiScale TRN & 34.44 & 33.60 & 48.80/77.64 & 50.85/79.33\\
      2-Stream TRN & 42.01 & 40.71 & 55.52/83.06 & 56.24/83.15  \\
      \bottomrule
      \end{tabular}
      
    }%
    \quad
    \subfloat[]{
    	\begin{tabular}{ p{0.7cm} p{0.9cm} p{0.8cm} }
        \toprule
         & TRN & TSN \\
        \hline
        2-fr. & 22.23 & 16.72 \\ 
        3-fr. & 26.22 & 17.30 \\
        5-fr. & 30.39 & 18.11 \\
        7-fr. & 31.01 & 18.48 \\

       \bottomrule
      \end{tabular}
    }%
    \caption{(a) Results on the validation set and test set of the Something-V1 Dataset (Top1 Accuracy) and Something-V2 Dataset (Both Top1 and Top5 accuracy are reported). (b) Comparison of TRN and TSN as the number of frames (fr.) varies on the validation set of the Something-V1. TRN outperforms TSN in a large margin as the number of frames increases, showing the importance of temporal order. }%
    \label{fig:something_result}%
    \vspace{-15mm}
\end{table}



\subsection{Results on Jester and Charades}

We further evaluate the TRN-equipped networks on the Jester dataset, which is a video dataset for hand gesture recognition with 27 classes. The results on the validation set of the Jester dataset are listed in Table \ref{fig:jester_result}a. The result on the test set and comparison with the top methods 
are listed in Table \ref{fig:jester_result}b. MultiScale TRN again achieves competitive performance as close to 95\% Top1 accuracy.

\begin{table}%
\vspace{-5mm}
    \centering
    \subfloat[]{
      \begin{tabular}{ p{2.4cm}  c  }
      \toprule                      
       & Val   \\
      \hline  
      Baseline & 63.60 \\
      2-frame TRN & 75.65 \\ 
      3-frame TRN & 81.45 \\
      4-frame TRN & 89.38 \\
      5-frame TRN & 91.40 \\ 
      \hline
      MultiScale TRN & \textbf{95.31} \\
      \bottomrule
      \end{tabular}
    }%
    \qquad
    \subfloat[]{
        \begin{tabular}{ l  c }
        \toprule                       
        & Test \\
        \hline   
        20BN Jester System & 82.34 \\
        VideoLSTM & 85.86 \\
        Guillaume Berger & 93.87  \\
        Ford's Gesture System & 94.11  \\
        Besnet & 94.23  \\
        \hline   
        MultiScale TRN & \textbf{94.78}  \\
        \bottomrule
        \end{tabular}
    }%
    \caption{Jester Dataset Results on (a) the validation set and (b) the test set.}%
    \label{fig:jester_result}%
    \vspace{-6mm}
\end{table}



We evaluate the MultiScale TRN on the recent Charades dataset for daily activity recognition. The results are listed in Table \ref{test_charades}. Our method outperforms various methods such as 2-stream networks and C3D \cite{sigurdsson2016hollywood}, and the recent Asynchronous Temporal Field (TempField) method \cite{sigurdsson2016asynchronous}. 

\begin{figure*}
\vspace{-10mm}
\centering
\includegraphics[width=0.9\linewidth]{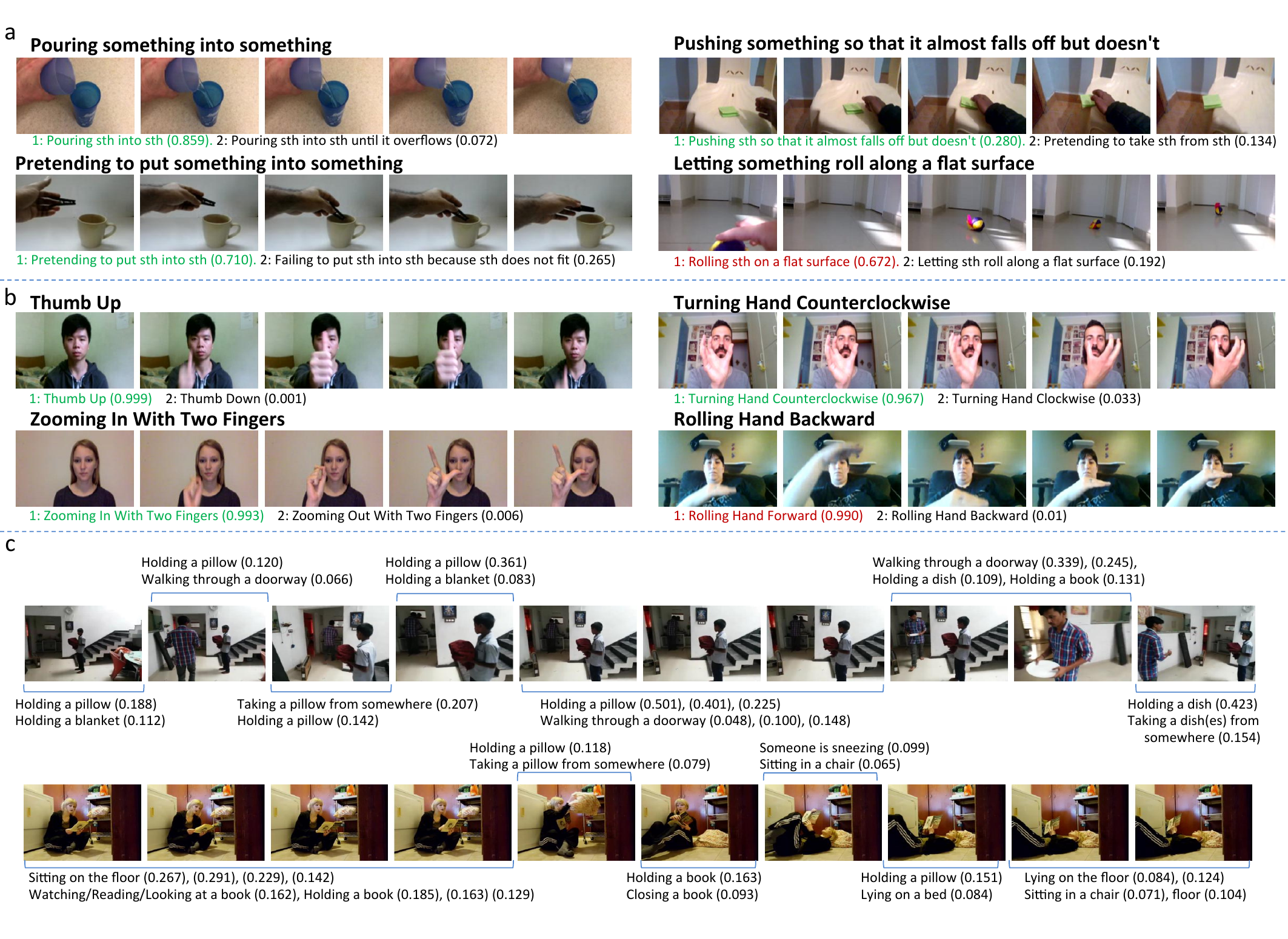}
\vspace{-5mm}
\caption{Prediction examples on a) Something-Something, b) Jester, and c)  Charades. For each example drawn from Something-Something and Jester, the top two predictions with green text indicating a correct prediction and red indicating an incorrect one. Top 2 predictions are shown above Charades frames.}
\vspace{-10mm}
\label{samples_prediction}
\end{figure*}

The qualitative prediction results of the Multi-Scale TRN on the three datasets are shown in Figure \ref{samples_prediction}. The examples in Figure \ref{samples_prediction} demonstrate that the TRN model is capable of correctly identifying actions for which the overall temporal ordering of frames is essential for a successful prediction. For example, the turning hand counterclockwise category would assume a different class label when shown in reverse. Moreover, the successful prediction of categories in which an individual \emph{pretends} to carry out an action (e.g. `pretending to put something into something' as shown in the second row) suggests that the network can capture temporal relations at multiple scales, where the ordering of several lower-level actions contained in short segments conveys crucial semantic information about the overall activity class.

This outstanding performance shows the effectiveness of the TRN for temporal relational reasoning and its strong generalization ability across different datasets.

\begin{table}
\vspace{-10mm}
\caption{Results on Charades Activity Classification.}
\label{test_charades}
\centering
\begin{tabular}{ l  c  c  c  c  c c c}
\toprule
Approach & Random & C3D & AlexNet & IDT & 2-Stream & TempField & Ours\\
\hline
mAP & 5.9 & 10.9 & 11.3 & 17.2 & 14.3 & 22.4 & \textbf{25.2} \\
\bottomrule
\end{tabular}
\vspace{-5mm}
\end{table}


\subsection{Interpreting Visual Common Sense Knowledge inside the TRN}

\begin{figure}
\vspace{-6mm}
\centering
\includegraphics[width=1\linewidth]{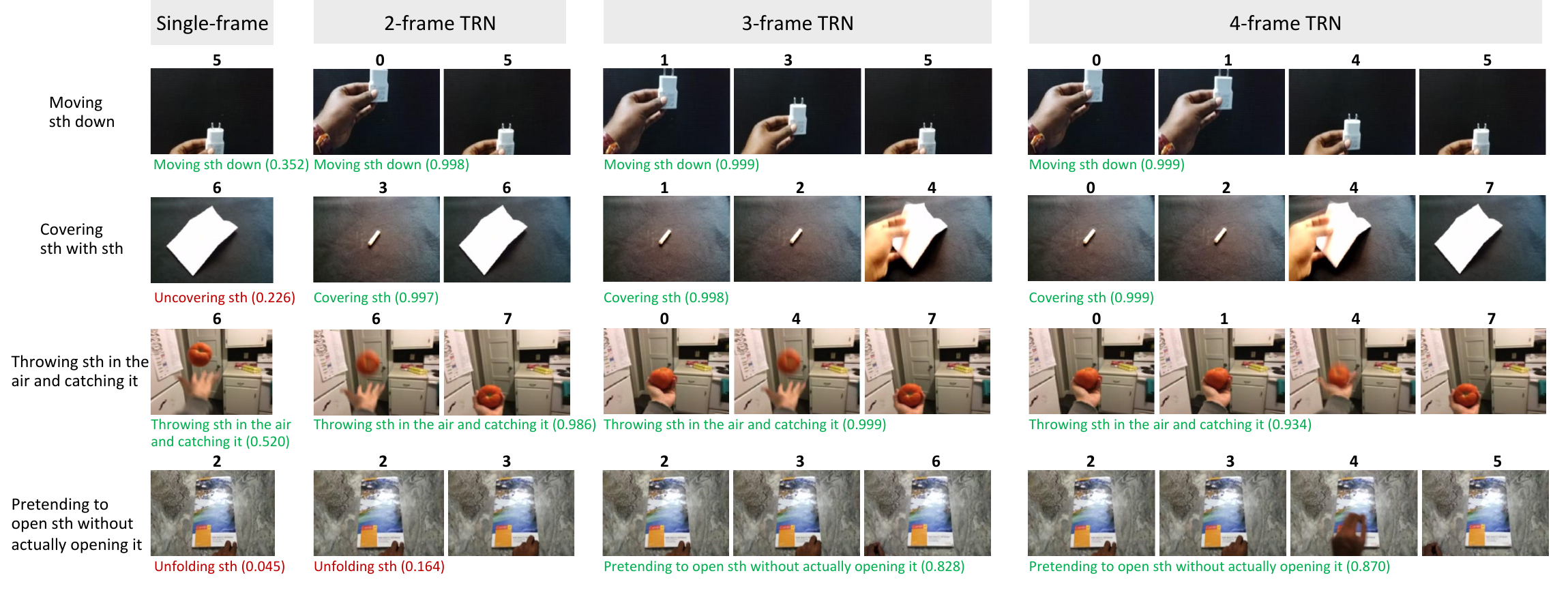}
\vspace{-7mm}
\caption{The top representative frames determined by single frame baseline network, the 2-frame TRN, 3-frame TRN, and 4-frame TRN. TRNs learn to capture the essence of an activity only given a limited number of frames. Videos are from the validation set of the Something-Something dataset}
\label{representative_frames}
\vspace{-4mm}
\end{figure}

One of the distinct properties of the proposed TRNs compared to previous video classification networks such as C3D \cite{tran2015learning} and I3D \cite{carreira2017quo} is that TRN has more interpretable structure. In this section, we have a more in-depth analysis to interpret the visual common sense knowledge learned by the TRNs through solving these temporal reasoning tasks. We explore the following four parts:

\textbf{Representative frames of a video voted by the TRN to recognize an activity}.  Intuitively, a human observer can capture the essence of an action by selecting a small collection of representative frames. Does the same hold true for models trained to recognize the activity? To obtain a sequence of representative frames for each TRN, we first compute the features of the equidistant frames from a video, then randomly combine them to generate different frame relation tuples and pass them into the TRNs. Finally we rank the relation tuples using the responses of different TRNs. Figure \ref{representative_frames} shows the top representative frames voted by different TRNs to recognize an activity in the same video. We can see that the TRNs learn the temporal relations that characterize an activity. For comparatively simple actions, a single frame is sufficient to establish some degree of confidence in the correct action, but is vulnerable to mistakes when a transformation is present. 2-frame TRN picks up the two frames that best describe the transformation. Meanwhile, for more difficult activity categories such as `Pretending to poke something', two frames are not sufficient information for even a human observer to differentiate. Similarly, the network needs additional frames in the TRNs to correctly recognize the behavior.

Thus the progression of representative frames and their corresponding class predictions inform us about how temporal relations may help the model reason about more complex behavior. One particular example is the last video in Figure \ref{representative_frames}:
The action's context given by a single frame - a hand close to a book - is enough to narrow down the top prediction to a qualitatively plausible action, unfolding something. A similar, two-frame relation marginally increases the probability the initial prediction, although these two frames would not be sufficient for even human observers to make the correct prediction.
Now, the three frame-relation begins to highlight a pattern characteristic to Something-Something’s set of \emph{pretending} categories: the initial frames closely resemble a certain action, but the later frames are inconsistent with the completion of that action as if it never happened. This relation helps the model to adjust its prediction to the correct class. Finally, the upward motion of the individual’s hand in the third frame of the 4-frame relation further increases the discordance between the \emph{anticipated} and \emph{observed} final state of the scene; a motion resembling the action appeared to take place with no effect on the object, thus, solidifying confidence in the correct class prediction.

\begin{figure}
\vspace{-10mm}
\centering
\includegraphics[width=0.9\linewidth]{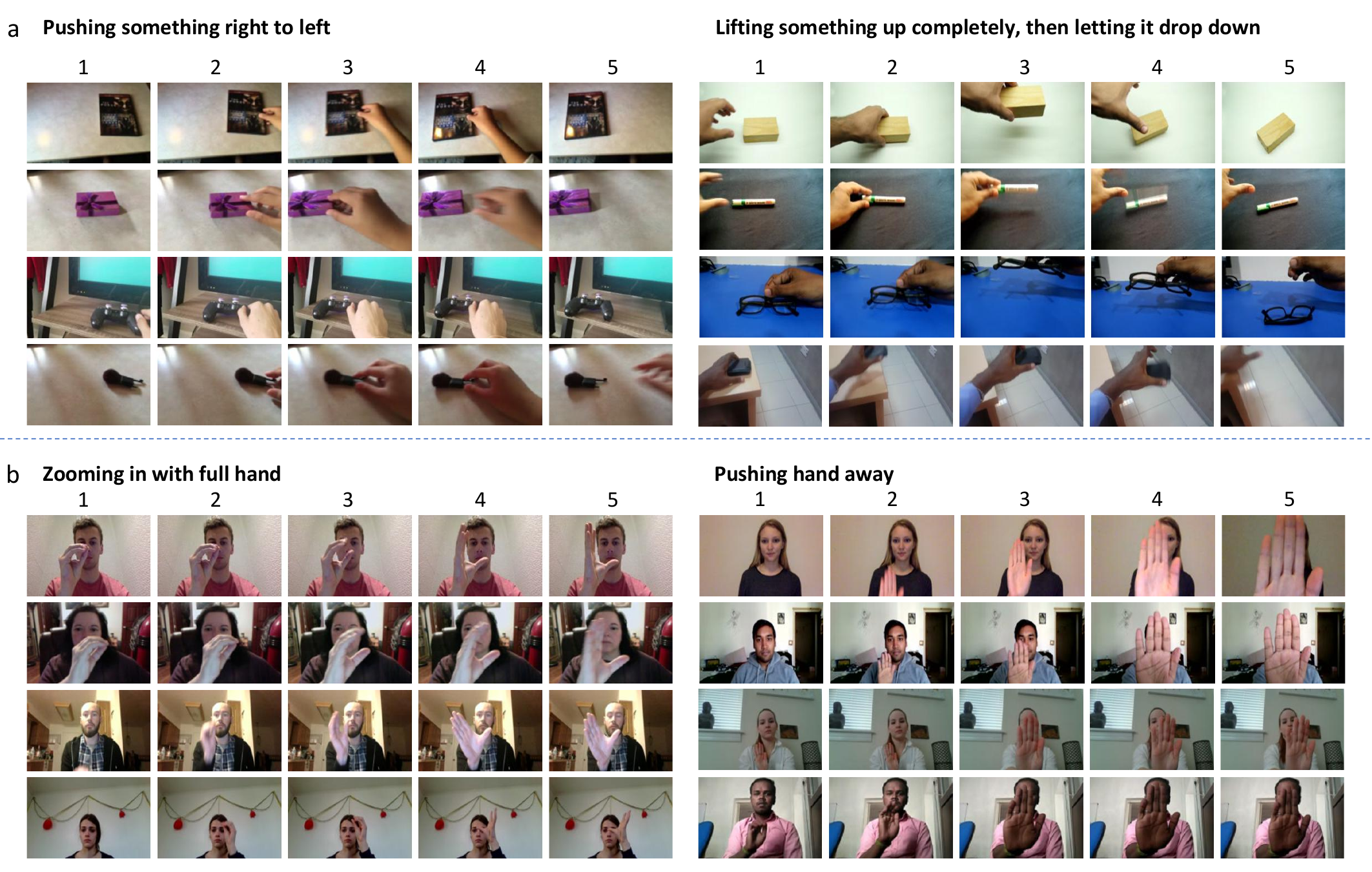}
\vspace{-3mm}
\caption{Temporal alignment of videos  from the (a) Something-Something and (b) Jester  datasets using the most representative frames as temporal anchor points. For each action, 4 different videos are aligned using 5 temporal anchor points.}
\label{video_alignment}
\vspace{-6mm}
\end{figure}

\textbf{Temporal Alignment of Videos}. The observation that the representative frames identified by the TRN are consistent across instances of an action category suggests that the TRN is well suited for the task of temporally aligning videos with one another. Here, we wish to synchronize actions across multiple videos by establishing a correspondence between their frame sequences. Given several video instances of the same action, we first select the most representative frames for each video and use their frame indices as ``landmark'', temporal anchor points.Then, we alter the frame rate of video segments between two consecutive anchor points such that all of the individual videos arrive at the anchor points at the same time. Fig.\ref{video_alignment} shows the samples from the aligned videos. We can see different stages of an action are captured by the temporal relation. The temporal alignment is also an exclusive application of our TRN model, which cannot be done by previous video networks 3D convNet or two-stream networks.

\begin{figure}%
\vspace{-10mm}
    \centering
    \subfloat[]{{\includegraphics[width=6cm]{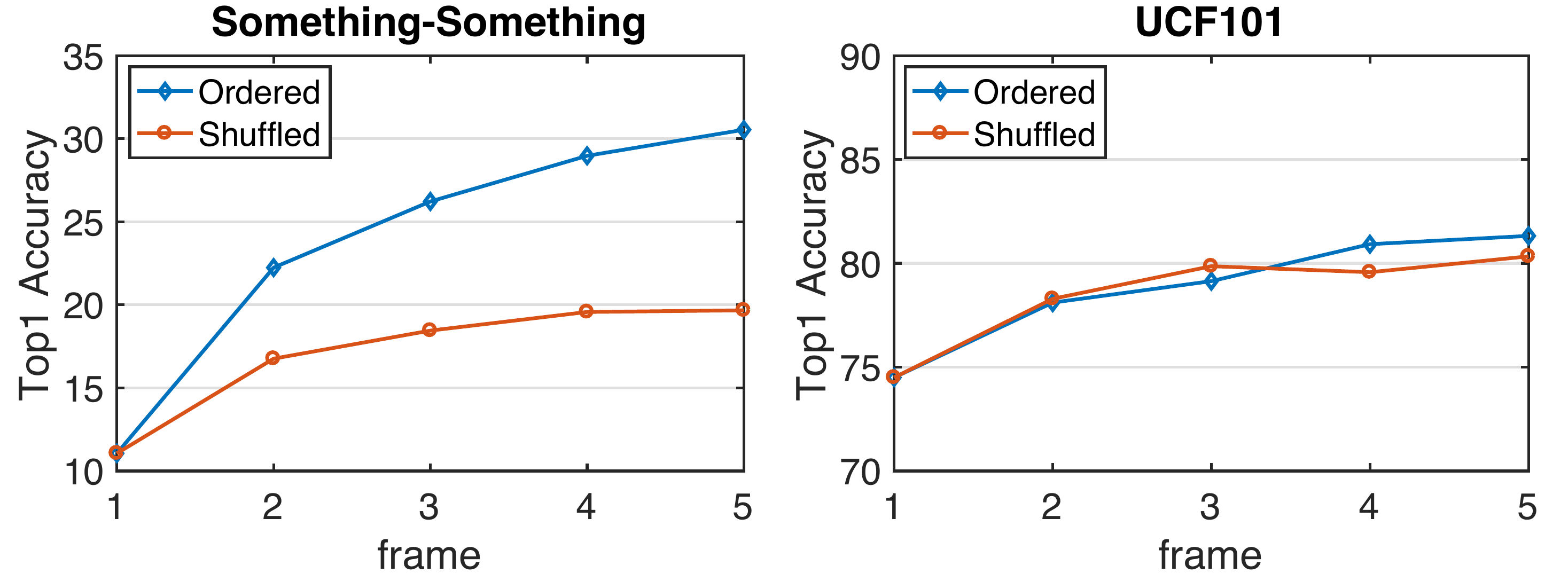} }}%
    \subfloat[]{{\includegraphics[width=5cm]{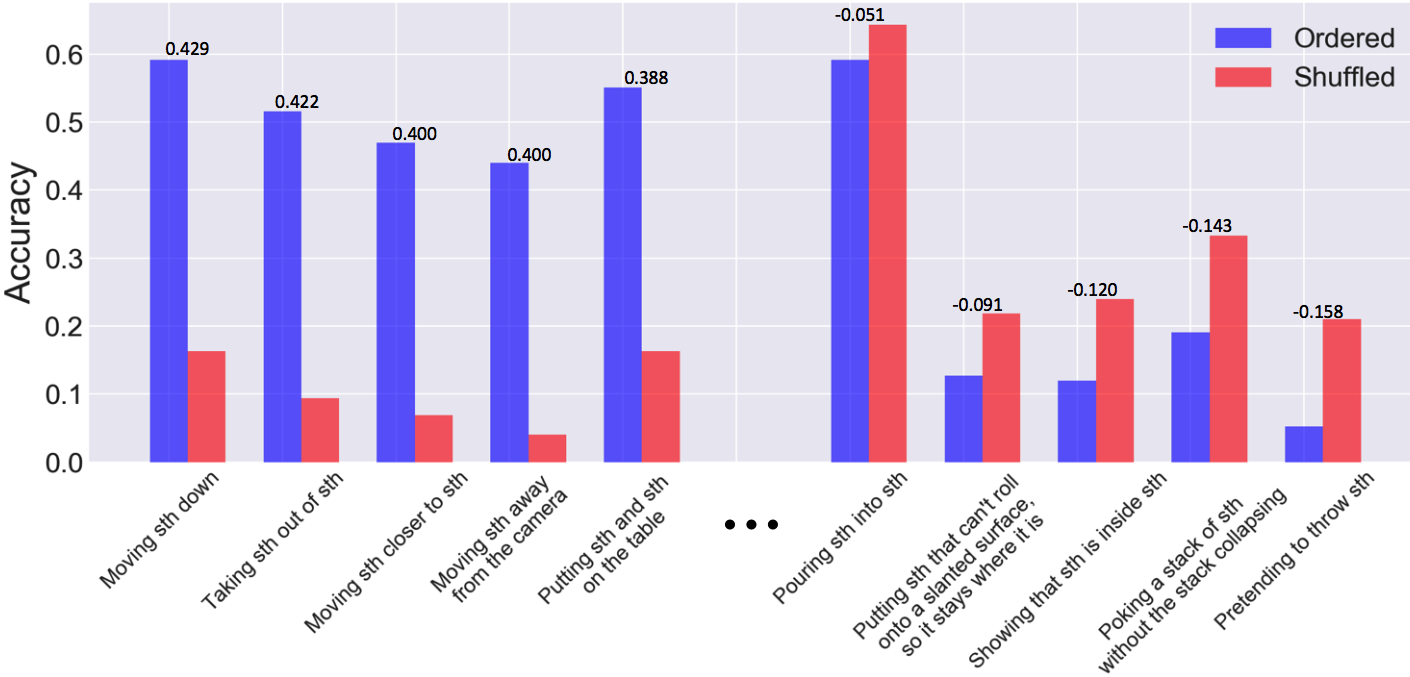} }}%
    \vspace{-4mm}
    \caption{(a) Accuracy obtained using ordered frames and shuffled frames, on Something-Something and UCF101 dataset respectively. On Something-Something, the temporal order is critical for recognizing the activity. But recognizing activities in UCF101 does not necessarily require temporal relational reasoning. (b) The top 5 action categories that exhibited the largest gain and the least gain (negative) respectively between ordered and shuffled frames as inputs. Actions with directional motion appear to suffer most from shuffled inputs.}
    \label{fig:frame_order}%
    \vspace{-3mm}
\end{figure}

\textbf{Importance of temporal order for activity recognition}. To verify the importance of the temporal order of frames for activity recognition, we conduct an experiment to compare the scenario with input frames in temporal order and in shuffled order when training the TRNs, as shown in Figure \ref{fig:frame_order}a. For training the shuffled TRNs, we randomly shuffle the frames in the relation modules. The significant difference on the Something-Something dataset shows the importance of the temporal order in the activity recognition. More interestingly, we repeat the same experiment on the UCF101 dataset \cite{soomro2012ucf101} and observe no difference between the ordered frames and shuffled frames. That shows activity recognition for the Youtube-type videos in UCF101 doesn't necessarily require the temporal reasoning ability since there are not so many casual relations associated with an already on-going activity.

To further investigate how temporal ordering influences activity recognition in TRN, we examine and plot the categories that show the largest differences in the class accuracy between ordered and shuffled inputs drawn from the Something-Something dataset, in Figure \ref{fig:frame_order}b. In general, actions with strong `directionality’ and large, one-way movements, such as `Moving something down', appear to benefit the most from preserving the correct temporal ordering. This observation aligns with the idea that the disruption of continuous motion and a potential consequence of shuffling video frames, would likely confuse a human observer, as it would go against our intuitive notions of physics.

Interestingly, the penalty for shuffling frames of relatively static actions is less severe if penalizing at all in some cases, with several categories marginally benefiting from shuffled inputs, as observed with the category `putting something that can't roll onto a slanted surface so it stays where it is'. Here, simply learning the coincidence of frames rather than temporal transformations may be sufficient for the model to differentiate between similar activities and make the correct prediction. Particularly in challenging ambiguous cases, for example `Pretending to throw something' where the release point is partially or completely obscured from view, disrupting a strong `sense of motion' may bias model predictions away from the likely alternative, `throwing something', frequently but incorrectly selected by the ordered model, thus giving rise to a curious difference in accuracy for that action.

The difference between TSN and TRN is at using different frame feature pooling strategies, where TRN using Temporal Relation(TR) pool emphasizes on capturing the temporal dependency of frames while TSN simply uses average pool to ignore the temporal order. We evaluate the two pool strategies in detail as shown in Table \ref{datasets_pool}. The difference in the performance using average pool and TR pool actually reflects the importance of temporal orders in a video dataset. The tested datasets are categorized by the video source, where the first three are Youtube videos, the other three are videos crowdsourced from AMT. The base CNN is BNInception. Both of the models use 8 frames. Interestingly, the models with average pool and TR pool achieve similar accuracy on Youtube videos, thus recognizing Youtube videos doesn't require much temporal order reasoning, which might be due to that activity in the randomly trimmed Youtube videos doesn't usually have a clear action start or end. On the other hand, the crowdsourced video has just one activity with clearly start and end, thus temporal relation pool brings significant improvement.

\begin{table}
\vspace{-6mm}
\begin{center}
\begin{tabular}{p{2.1cm} p{1.4cm} p{1.4cm} p{1.5cm} p{0.1cm} p{1.7cm}  p{1.3cm} p{1.8cm}}
 \hline
 & \multicolumn{3}{c}{Youtube videos} & & \multicolumn{3}{c}{Crowdsourced videos} \\ \cmidrule{2-4} \cmidrule{6-8} 
 Dataset & UCF & Kinetics & Moments & & Something & Jester & Charades\\
 \hline
 Num.Classes & 101 & 200 & 339 & & 174 & 27 & 157 \\
 \hline
Average Pool & 82.69 & \textbf{63.34} &  24.11 & & 19.53 & 85.41 & 11.32  \\
TR Pool & \textbf{83.83} & 63.18 & \textbf{25.94} &  & \textbf{34.44} &  \textbf{95.31} & \textbf{25.20}\\
\hline
\end{tabular}
\caption{Accuracy on six video datasets for models with two pool strategies.}
\label{datasets_pool}
\end{center}
\vspace{-12mm}
\end{table}

\begin{figure}
\vspace{-10mm}
\centering
\includegraphics[width=0.90\linewidth]{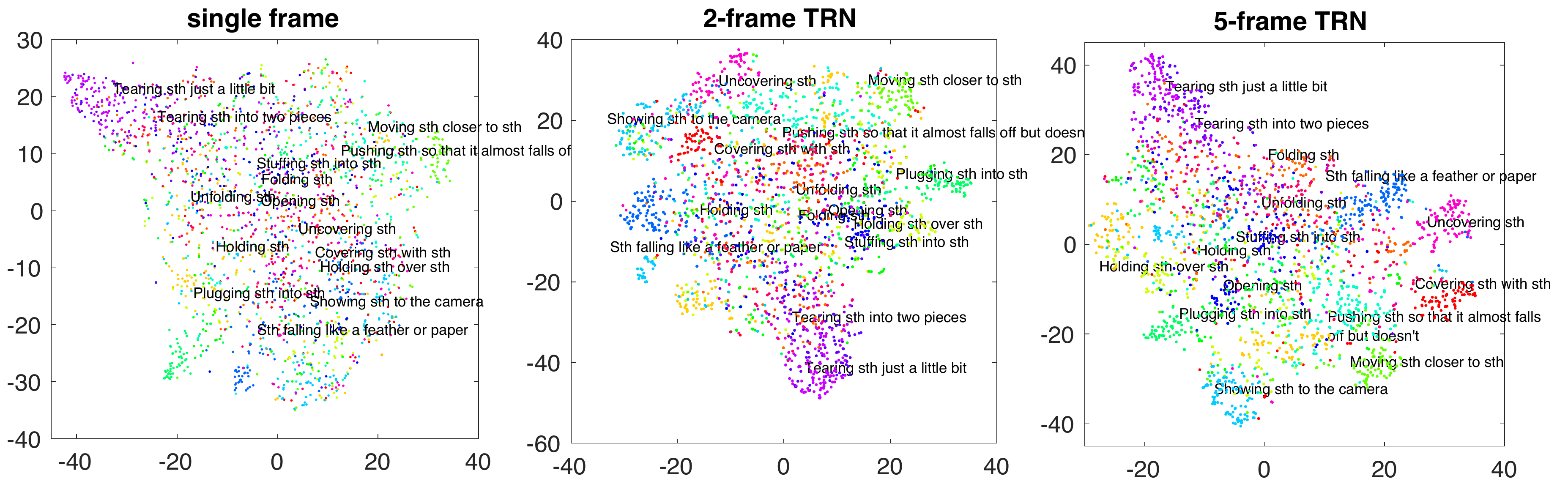}
\vspace{-4mm}
\caption{t-SNE plot of the video samples of the 15 classes using the deep features from the single-frame baseline, 2-frame TRN, and 5-frame TRN. Higher frame TRN can better differentiate activities in Something-Something dataset.}
\vspace{-5mm}
\label{tsne}
\end{figure}

\textbf{t-SNE visualization of activity similarity}. Figure \ref{tsne} shows the t-SNE visualization for embedding the high-level features from the single frame baseline, the 3-frame TRN, and the 5-frame TRN, for the videos of the 15 most frequent activity classes in the validation set. We can see that the features from 2-frame and 5-frame TRNs can better differentiate activity categories. We also observe the similarity among categories in the visualization map. For example, `Tearing something into two pieces' is very similar to `Tearing something just a little bit', and the categories `Folding something', `Unfolding something', `Holding something', `Holding something over something' are clustered together. 

\begin{table}
\vspace{-6mm}
\caption{Early activity recognition using the MultiScale TRN on Something-Something and Jester dataset. Only the first 25\% and 50\% of frames are given to the TRN to predict activities. Baseline is the model trained on single frames. }\label{forecasting}
\small
\centering
\begin{tabular}{ p{1.5cm}  c c p{0.3cm}  c  c}
\toprule 
& \multicolumn{2}{c}{Something} & & \multicolumn{2}{c}{Jester} \\
\cline{2-3} \cline{5-6}
Frames & baseline & TRN & & baseline & TRN \\
\hline   
first 25\% & 9.08 & 11.14 & & 27.25 & 34.23 \\
first 50\% & 10.10 & 19.10 & & 41.43 & 78.42 \\
full & 11.41 & 33.01 & & 63.60 & 93.70 \\
\bottomrule
\end{tabular}
\vspace{-4mm}
\end{table}

\begin{figure}
\centering
\includegraphics[width=0.9\linewidth]{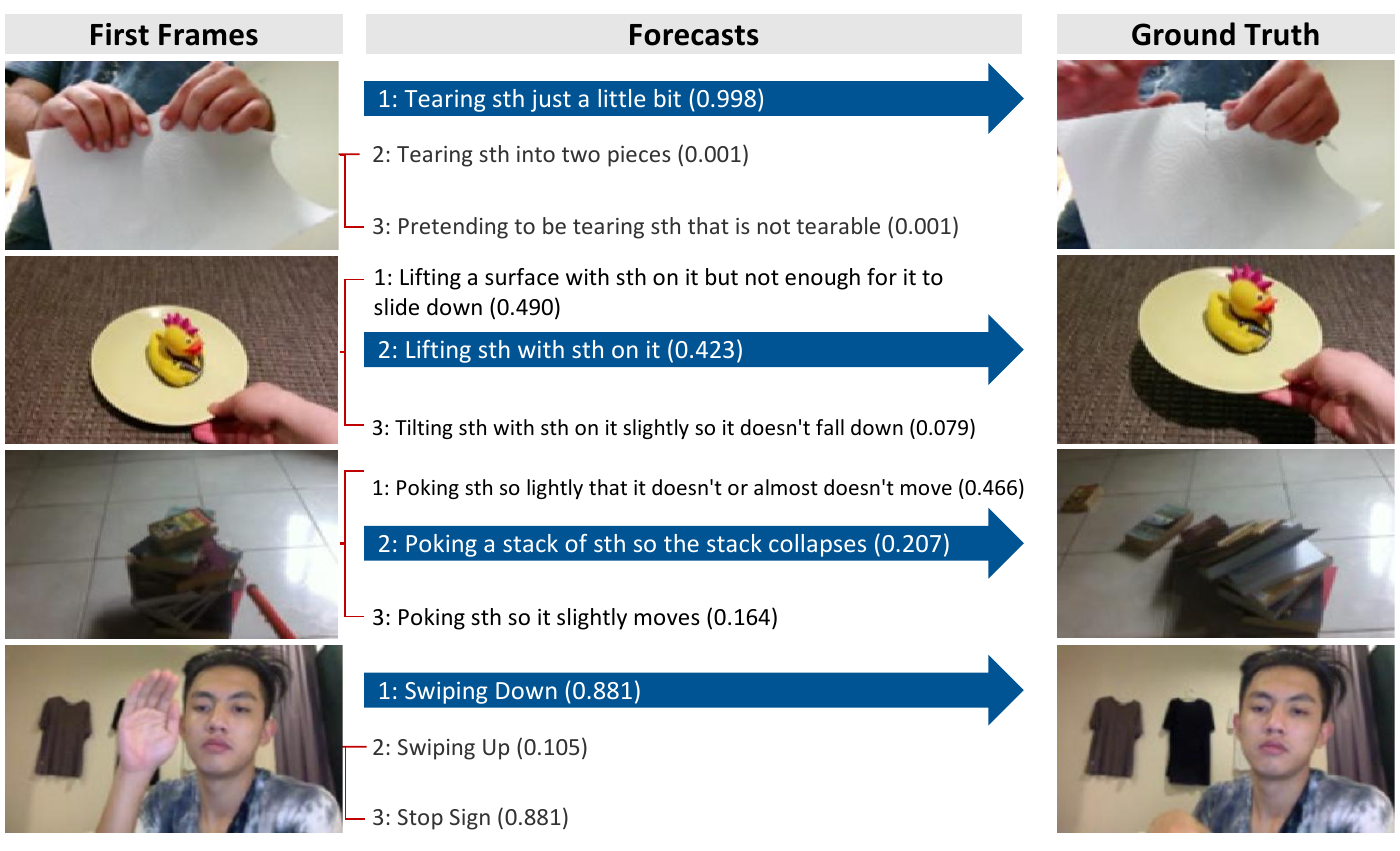}
\vspace{-3mm}
\caption{Early recognition of activity when only given the first 25\% frames. The first 25\% of each video, represented by the first frame shown in the left column, is used to generate the top 3 anticipated forecasts and corresponding probabilities listed in the middle column. The ground truth label is highlighted by a blue arrow which points to the last frame of the video on the right.}
\label{forecasting_qualitative}
\vspace{-4mm}
\end{figure}

\textbf{Early Activity Recognition}. Recognizing activities early or even anticipating and forecasting activities before they happen or fully happen is a challenging yet less explored problem in activity recognition. Here we evaluate our TRN model on early recognition of activity when given only the first 25\% and 50\% of the frames in each validation video. Results are shown in Table \ref{forecasting}. For comparison, we also include the single frame baseline, which is trained on randomly sampled individual frames from a video. We see that TRN can use the learned temporal relations to anticipate activity. The performance increases as more ordered frames are received. Figure \ref{forecasting_qualitative} shows some examples of anticipating activities using only first 25\% and 50\% frames of a video. A qualitative review of these examples reveals that model predictions on only initial frames do serve as very reasonable forecasts despite being given task with a high degree of uncertainty even for human observers. 

\section{Conclusion}
We proposed a simple and interpretable network module called Temporal Relation Network (TRN) to enable temporal relational reasoning in neural networks for videos. We evaluated the proposed TRN on several recent datasets and established competitive results using only discrete frames. Finally, we have shown that TRN modules discover visual common sense knowledge in videos.

\tiny
\textbf{Acknowledgement}: This work was partially funded by DARPA XAI program No. FA8750-18-C-0004, NSF Grant No. 1524817, and Samsung to A.T.; the Vannevar Bush Faculty Fellowship program funded by the ONR grant No. N00014-16-1-3116 to A.O.. It is also supported in part by the Intelligence Advanced Research Projects Activity (IARPA) via Department of Interior/ Interior Business Center (DOI/IBC) contract number D17PC00341. The U.S. Government is authorized to reproduce and distribute reprints for Governmental purposes notwithstanding any copyright annotation thereon. Disclaimer: The views and conclusions contained herein are those of the authors and should not be interpreted as necessarily representing the official policies or endorsements, either expressed or implied, of IARPA, DOI/IBC, or the U.S. Government.
\clearpage

\bibliographystyle{splncs}
\bibliography{egbib}
\end{document}